\documentclass{article}

\usepackage{PRIMEarxiv}

\usepackage[utf8]{inputenc} % allow utf-8 input
\usepackage[T1]{fontenc}    % use 8-bit T1 fonts
\usepackage{hyperref}       % hyperlinks
\usepackage{url}            % simple URL typesetting
\usepackage{booktabs}       % professional-quality tables
\usepackage{amsfonts}       % blackboard math symbols
\usepackage{nicefrac}       % compact symbols for 1/2, etc.
\usepackage{microtype}      % microtypography
\usepackage{lipsum}
\usepackage{fancyhdr}       % header
\usepackage{graphicx}       % graphics
\graphicspath{{media/}}     % organize your images and other figures under media/ folder
\usepackage{algorithmic}
\usepackage{algorithm}
\usepackage{multicol,multirow}
\usepackage{amsmath,amssymb,amsfonts}
\usepackage{authblk}
\title{TROI: Cross-Subject Pretraining with Sparse Voxel Selection for Enhanced fMRI Visual Decoding}

\author[1]{Ziyu Wang, Tengyu Pan, Zhenyu Li, Jianyong Wang}
\author[2]{Xiuxing Li}
\author[3]{Ji Wu}
\affil[1]{ Department of Computer Science and Technology, Tsinghua University, Beijing, China}
\affil[2]{ School of Computer Science and Technology, Beijing Institute of Technology , Beijing, China}
\affil[3]{ School of Computer Science and Engineering, Beihang University, Beijing, China}

\begin{document}
\maketitle

\begin{abstract}
fMRI (functional Magnetic Resonance Imaging)  visual decoding involves decoding the original image from brain signals elicited by visual stimuli. This often relies on manually labeled ROIs (Regions of Interest) to select brain voxels.
However, these ROIs can contain redundant information and noise, reducing decoding performance. Additionally, the lack of automated ROI labeling methods hinders the practical application of fMRI visual decoding technology, especially for new subjects.
This work presents TROI (Trainable Region of Interest), a novel two-stage, data-driven ROI labeling method for cross-subject fMRI decoding tasks, particularly when subject samples are limited.
TROI leverages labeled ROIs in the dataset to pretrain an image decoding backbone on a cross-subject dataset, enabling efficient optimization of the input layer for new subjects without retraining the entire model from scratch. In the first stage, we introduce a voxel selection method that combines sparse mask training and low-pass filtering to quickly generate the voxel mask and determine input layer dimensions. In the second stage, we apply a learning rate rewinding strategy to fine-tune the input layer for downstream tasks.
Experimental results on the same small sample dataset as the baseline method for brain visual retrieval and reconstruction tasks show that our voxel selection method surpasses the state-of-the-art method MindEye2 with an annotated ROI mask.
\end{abstract}

% keywords can be removed

\section{Introduction}
\label{sec:intro}
\begin{figure}[ht]
\centering
    \includegraphics[width=0.6\linewidth]{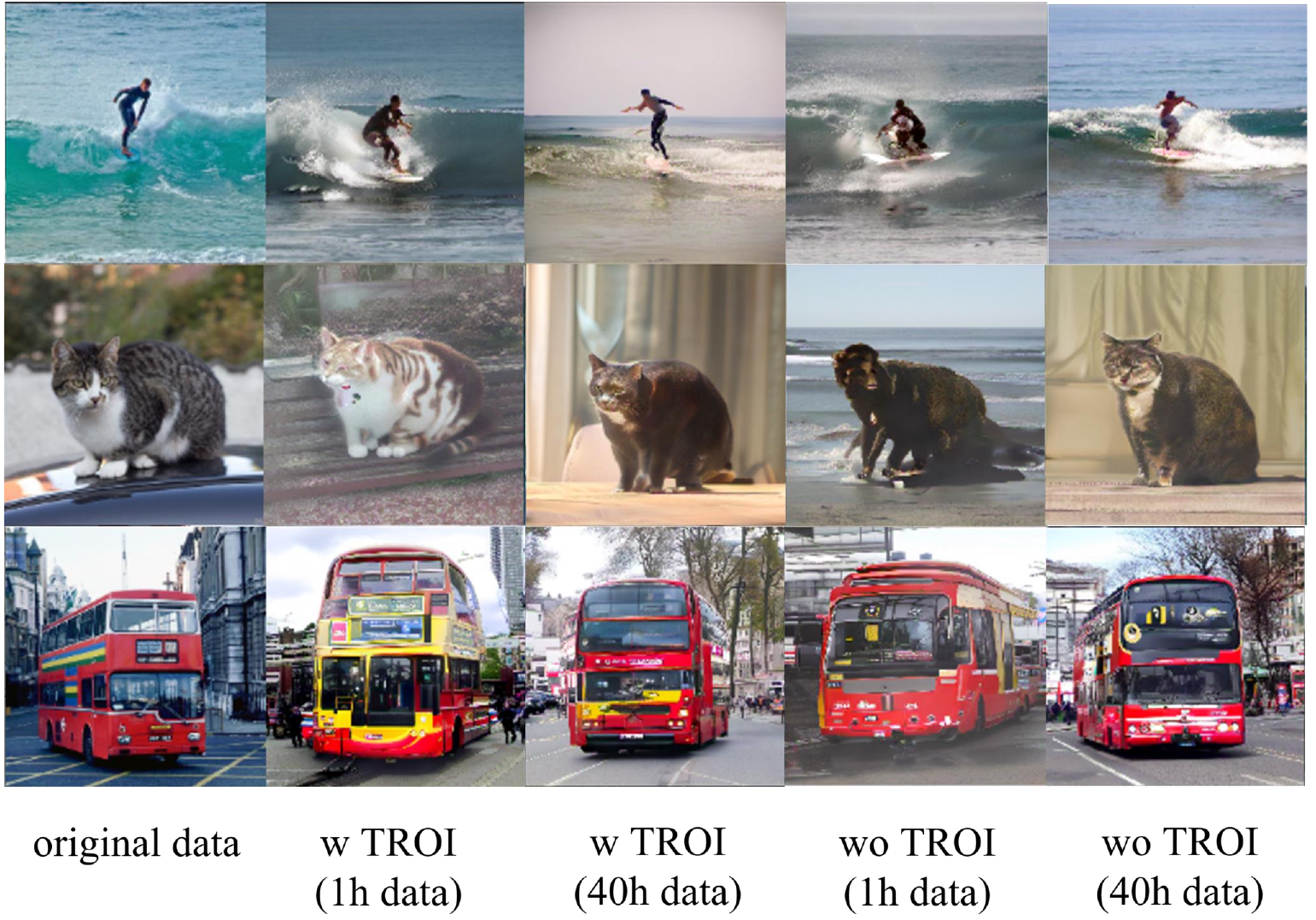}
    \caption{Visual reconstructions from fMRI using baseline MindEye2 w/wo our proposed TROI in different training data volume.}
    \label{fig:recon}
    % \vspace{-0.5cm}
\end{figure}

Decoding visual stimuli from measured brain activity has long been a central focus in computational neuroscience. Recently, significant progress has been made in decoding techniques that focus on functional magnetic resonance imaging (fMRI) data. fMRI is a type of neuroimaging that reflects neural activity by measuring changes in brain oxygenation levels \cite{huang2021fmri}.
Recent methods\cite{ME1,ME2,ozcelik2022reconstruction,ozcelik2023natural,kneeland2023reconstructing} attempt to establish a mapping relationship between fMRI brain activity patterns and the latent space of pretrained deep learning models, for the purpose of retrieving and reconstructing visual stimuli corresponding to brain activity, as Fig. \ref{fig:recon} shows. 
% Anatomical alignment methods, such as the Talairach \cite{dervin1990co} or MNI templates \cite{mazziotta2001probabilistic}, are often used to standardize brains across the subjects. 
% However, these methods can distort brain shape and disrupt the spatial information in fMRI data, negatively affecting decoding performance.

While these methods have demonstrated notable success, they encounter limitations when applied to new subjects due to individual variations in brain shape and functional organization. To mitigate this issue, existing approaches often rely on regions of interest (ROIs) to perform decoding, which helps reduce the impact of structural and functional differences across subjects. Common techniques include anatomical template-based ROI analysis \cite{talairach1988co, mazziotta2001probabilistic}, functional-based ROI analysis \cite{van2013wu, arcaro2015anatomical, NSD}, and data-driven ROI analysis \cite{beckmann2005investigations, yeo2011organization}. However, these methods either neglect subject-specific variations or focus exclusively on regions associated with simple visual stimuli. For more complex decoding tasks, such as image reconstruction, the selected ROIs may contain redundant and noisy voxels, and the large encoding space of task events can lead to suboptimal performance.
% However, prior methods are either based on predefined templates, which overlook individual subject differences, or can only identify stimulus-related regions. These methods fail in high-dimensional decoding tasks, such as reconstruction, due to the large encoding space of task events.

% The lack of annotated ROIs and the limited training samples for new subjects present significant challenges in extending brain decoding models.
To address these issues, we propose a two-stage data-driven ROI labeling method called Trainable Region of Interest(TROI).
This method is for cross-subject fMRI decoding tasks, even with limited samples from a new subject. 
Before labeling for the new subject, we pretrain an fMRI decoding backbone on a cross-subject dataset to focus on optimizing the brain voxel mask rather than training the entire model from scratch.
In the first stage, we apply LASSO regularization\cite{tibshirani1996regression} for sparse mask training and use a low-pass filter to quickly obtain the voxel mask and determine the input layer dimensions. In the second stage, we fine-tune the input layer for downstream tasks using a learning rate rewinding strategy.
When evaluated on small samples from the NSD dataset using the state-of-the-art fMRI decoding model, MindEye2, our approach outperforms methods with annotated ROIs in both retrieval and reconstruction tasks.

% 具体而言，我们的创新点有如下两条：

% 我们的方法在哪个任务上surpass了哪个baseline

% novelty 是什么！ indetail! besides, 做了什么，有什么novelty!

% 与传统的压缩感知（CS）算法不同，我们的大脑体素掩码训练方法并非以在fMRI单模态空间中寻找表示稀疏解为目的，而是受跨模态图像信息的监督，其次，我们的方法不改变大脑体素的原始空间，这使得我们可以运用大脑区域的空间局部性特征来优化我们训练得到的掩码。（这里可以compare）
% 我们方法的好处之二在于，在rewind期间可以使用更大的batchsize进行训练

% Additionally, when handling small sample learning, we adopted a \textbf{pretraining and fine-tuning learning mechanism}
% %这个预训练-微调不一定对%
% with the manifold mix-up data augmentation method to expand the dataset, addressing the issue of insufficient training data.

\begin{figure*}[ht]
\centering
    \includegraphics[width=1.0\linewidth]{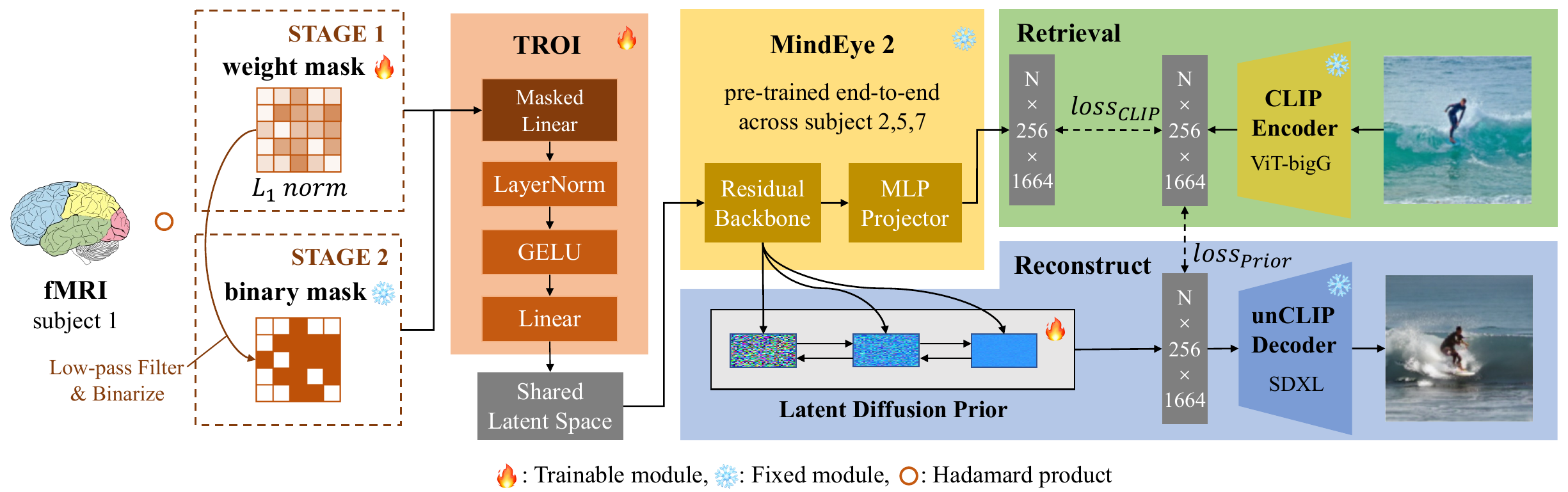}
    \caption{Overview of TROI module structure followed by a residual MLP backbone and two task-specific submodules. Taking Subject 1 as a new subject, the cross-subject backbone network is trained using data from Subjects 2, 3, 5, and 7. The subject-specific TROI module is then trained in two stages with small sample data from Subject 1.}
    \label{fig:model structure}
    % \vspace{-0.5cm}
\end{figure*}

\section{Approach}
As shown in Fig. \ref{fig:model structure}, our model maps brain activity into an embedding space using the subject-specific TROI and subject-shared residual Multiple Layer Perceptron (MLP) backbone networks. It then integrates various downstream tasks, such as retrieval and reconstruction, to facilitate cross-modal learning with image-based supervision signals. We begin by pretraining the backbone using data from multiple subjects, excluding one subject as a reserved subject. Then, we fine-tune the model using a subset of data from the reserved subject to simulate handling limited data from a new subject.
\label{sec:format}
\subsection{Cross-subject pretraining}
% 加点儿填充物

Denote the baseline network as $\mathcal{B}$,
a pretrained CLIP image encoder as $\mathcal{C}$.
Suppose we have $n$ subjects, and the dataset from $i$-th subject are $\mathcal{D}_i={\{x_{i,j},y_{i,j}\}}$, where $j$ refers to the $j$-th data point, $x_{i,j}$ and $y_{i,j}$ refer to fMRI and image sample of $i$-th subject and $j$-th data point respectively.
We flatten the spatial patterns of $x_{i,\cdot}$ and then align them to a shared latent space through subject-specific networks named as $TROI_i$ with the parameters $w_i$, to address the varying fMRI size of different subjects. 
We use MixCo\cite{kim2020mixco}, a linear interpolation method, to augment the data.
For subject $i$, suppose the sample size is $s_i$. We mix the fMRI data $x_{i,j}$ of the same subject $i$ but different data point $j$, as shown in Eq. (\ref{mixup}).
\begin{equation}
\label{mixup}
% \begin{aligned}
    x^*_{i,j} = \lambda_{i,j} x_{i,j}+(1-\lambda_{i,j}) x_{i,k_j}, \lambda \sim Beta(\alpha,\beta).
% \end{aligned}
\end{equation}
Here $k_j $ is an arbitrary mixing index for the $j$-th data, and $\lambda_{i,j}$ represents the $j$-th mixing weight, which is sampled from a Beta distribution with hyperparameter $\alpha=\beta=0.2$. 
The embedding of fMRI and image is calculated as $b^*_{i,j} = \mathcal{B}(TROI_i(x^*_{i,j}))$ and $c_{i,j} = \mathcal{C}(y_{i,j})$ respectively.

To address the challenge of learning an embedding space that can satisfy the objectives of both tasks \cite{ME1}, we incorporate an MLP Projector module $\mathcal{H}$ to decouple the feature spaces for retrieval and reconstruction. The projected embedding of fMRI is calculated as $h^*_{i,j} = \mathcal{H}(b^*_{i,j})$.

% Accordingly, the $b_{i,j}$ and $h_{i,j}$ with mixing are computed as follows.
% \begin{equation}
%     \begin{aligned}
%             b^*_{i,j} =& \mathcal{B}(TROI_i(x^*_{i,j})),\\
%     h^*_{i,j} =& \mathcal{H}(b^*_{i,j}).
%     \end{aligned}
% \end{equation}
\subsubsection{Retrieval}

% 在检索任务中，我们希望同一个批次中fMRI嵌入与与之对应的图像嵌入的余弦相似度最高，即：
In the retrieval task, our goal is to retrieve the corresponding image from a set of images based on a given fMRI data, and vice versa.
% To achieve this, we maximize the cosine similarity between the fMRI embedding and its corresponding image embedding within the same batch. 
To achieve this, we use the CLIP loss with MixCo to maximize the cosine similarity between matching pairs and minimize it for non-matching pairs, as shown in Eq. (\ref{CLIP loss}).
% eq 5 simplify
% \begin{equation}
% \begin{aligned}
% \label{CLIP loss}
% \mathcal{L}_{CLIP} =& -\sum_{i=1}^n\sum_{j=1}^{s_i}
% \lambda_{i,j} \log\left(\frac{\exp\left(\frac{h^*_{i,j}\cdot c_{i,j}}{\tau}\right)}{\sum^n_{p=1}\sum^{s_p}_{q=1}\exp\left(\frac{h^*_{i,j}\cdot c_{p,q}}{\tau}\right)}\right) \\
% +&(1-\lambda_{i,j}) \log\left(\frac{\exp\left(\frac{h^*_{i,j}\cdot c_{i,k_j}}{\tau}\right)}{\sum^n_{p=1}\sum^{s_p}_{q=1}\exp\left(\frac{h^*_{i,j}\cdot c_{p,q}}{\tau}\right)}\right).
% \end{aligned}
% \end{equation}
\begin{equation}
    \begin{aligned}
\label{CLIP loss}
    \mathcal{L}_{CLIP} = \mathcal{L}_{MixCo}(h^*_{i,j},c_{i,j})
    \end{aligned}
\end{equation}
Here, $\mathcal{L}_{MixCo}$ is the loss function introduced in study \cite{kim2020mixco}, $h^*_{i,j}$ and $c_{i,j}$ are l2-normalized and $\tau$ is a temperature hyperparameter.
The study in \cite{liu2023over} mentions that mix-up augment can negatively impact the performance of generative tasks. To balance the performance of the two downstream tasks, we stop using the mix-up data augmentation method after one-third of the training process.

\subsubsection{Reconstruction}
% 这里介绍重建的模型结构、为什么需要prior、以及prior loss
In the reconstruction task, our goal is to reconstruct the image corresponding to what the subject saw based on the given fMRI data.
Following DALL-E 2 \cite{Ramesh}, we use a diffusion prior to map the fMRI embedding space to the image embedding space. Afterward, we use an unCLIP decoder to reconstruct the images.

To maintain the consistency between the reconstructed image and the original image, we minimize the Euclidean distance between the fMRI embedding vector and the corresponding image embedding.
Denote latent diffusion prior network as $\mathcal{P}$ with the parameters $\Omega$,
the loss of reconstruction is calculated as follows.
\begin{equation}
    \begin{aligned}
        \mathcal{L}_{Prior} =& \sum^n_{i=1}\sum^{s_i}_{j=1}(\mathcal{P}(b^*_{i,j})-(\lambda c_{i,j}+(1-\lambda)c_{i,k_j}))^2,
    \end{aligned}
\end{equation}
Here, $k_j$ is an arbitrary mixing index for the j-th data, consistent with the definition in Eq. (\ref{mixup}).
% and then we use a pretrained unCLIP Decoder transfer this embedding to a natual image.

During training, we optimize the model parameters $\mathcal{M},w,\Omega$ jointly by minimizing the losses from both tasks, balanced by a hyperparameter $\varepsilon$, as shown in Eq. (\ref{loss}).
\begin{equation}
\label{loss}
    \mathcal{L}(\mathcal{M},w,\Omega;\mathcal{B},\mathcal{H},\mathcal{C},\mathcal{D}) = \mathcal{L}_{CLIP} + \varepsilon \mathcal{L}_{Prior},
\end{equation}

% \subsection{Small Sample Fine Tuning}
% % 我们的小样本微调使用预训练阶段得到的backbone网络参数为一个新的受试者训练TROI模型和下游模型，分为两个部分：stage1是对TROI模块进行稀疏训练以训练得到一个大脑体素掩码，stage2则是使用该掩码训练TROI模块和下游任务模块。
% % 我们的方法可以处理缺失ROI信息的新受试者数据集，并且相较于原数据集中提供的ROI信息，我们的方法可以在检索和重建方面取得更优的结果
% % 我们将mindeye2中fMRI 骨干网络的输入层用TROI模块替换以进行ROI训练
% Our small-sample fine-tuning process is divided into two stages: Stage 1 involves sparsely training the TROI module to generate a brain voxel mask, while Stage 2 uses this mask to train both the TROI module and the downstream task modules. 

\subsection{Stage1: Sparse Mask Training}
% 在stage1中，我们试图使用预训练阶段得到的backbone网络参数为一个新的受试者训练TROI模型和下游模型。在训练时，为了避免复杂模型在小样本的情况下过拟合，我们固定backbone网络的参数，仅梯度更新TROI模型的参数。为了加速
% 当数据集中缺失ROI数据时，直接将所有大脑体素作为输入以训练模型会因为引入了大量噪声而使得模型出现过拟合的现象，这一问题在样本量有限的情况下更加严重。因此面对缺失ROI数据的小样本训练集，我们首先
% In Stage 1, we fix the pretrained backbone and trying to select the task-relative voxels from a new subject's fMRI data without annotated ROI. 
% Unlike traditional compressed sensing (CS) algorithms, our brain voxel mask training method does not aim to find sparse representations in the single-modality fMRI space. Instead, it is guided by supervision from cross-modal image information. 
Utilizing all brain voxels as input for model training introduces considerable noise, which can lead to overfitting, particularly in scenarios with limited sample sizes. To mitigate this issue, we employ a 0-1 ROI mask $\mathcal{M'}$ to selectively extract voxels from coarsely segmented brain regions, with a primary focus on the occipital lobe \cite{grill2004human}. This mask is applied to the input data through an element-wise multiplication (Hadamard product), resulting in a new input representation  $\hat{x}_i = x_i \circ \mathcal{M'}$.

When given a voxel number budget \( V \), the problem of determining the number of filters can be formulated as a mixed 0-1 binary optimization problem. The objective is to select a set of voxels that minimizes the decoding task loss function, while ensuring that the number of voxels in the mask does not exceed \( V \), as shown in Eq. (\ref{l0norm}).
% This formulation balances the trade-off between selecting the most informative voxels and adhering to the voxel budget, 
\begin{equation}
\begin{aligned}
\label{l0norm}
    \mathop{\min}_{\mathcal{M'},w,\Omega}& \mathcal{L}(\mathcal{M'},w,\Omega;\mathcal{B},\mathcal{H},\mathcal{C},\mathcal{D}),\\
s.t.& \Vert\mathcal{M'}\Vert_0\leq V, 
\end{aligned}
\end{equation}
However, since the loss function in Eq. (\ref{l0norm}) cannot be directly optimized via gradient-based methods, we introduce a weighted ROI mask, denoted as 
$\mathcal{M}$.  The loss function for the optimization problem is formulated by introducing an LASSO coefficient, denoted as \(\psi\), to adjust the optimization process.
\begin{equation}
\label{total loss}
    \mathcal{L}_{total} = \mathcal{L}(\mathcal{M},w,\Omega;\mathcal{B},\mathcal{H},\mathcal{C},\mathcal{D})+\psi\Vert\mathcal{M}\Vert_1,
\end{equation}

To accelerate the sparsification of $\mathcal{M}$, in each iteration, we set the elements of $\mathcal{M}$ that are less than a fixed threshold $th$ to zero.
$\mathcal{M'}$ can be calculated from $\mathcal{M}$, as shown in Eq. (\ref{binary}).
\begin{equation}
\label{binary}
    \mathcal{M'}[i]=\begin{cases}
    1& if \ \mathcal{M}[i] >th \\ 
    0& otherwise
\end{cases}
\end{equation}

In scenarios with limited training data, the weighted mask \(\mathcal{M}\) may learn incorrect weights due to overfitting. In such cases, directly applying a fixed threshold for binarization can result in a voxel mask \(\mathcal{M}'\) that includes incorrect voxels, leading to poor performance. 
To address this issue, we apply a low-pass filter \(G\) to \(\mathcal{M}\) before binarization. This step leverages the spatial locality of brain functional regions, incorporating prior information from the data to mitigate overfitting. The proposed algorithm proceeds iteratively, as shown in Algorithm \ref{algo:Sparse mask training}.

\begin{algorithm}
\caption{Sparse Mask Training}
\label{algo:Sparse mask training}
\begin{algorithmic}[1]

\REQUIRE Pretrained backbone $\mathcal{B}$ and projector $\mathcal{H}$,
Pretrained CLIP image encoder $\mathcal{C}$,
Small sample training dataset $\mathcal{D}$\\
\STATE initialize $\mathcal{M}$ and $\mathcal{M'}$ as all ones

\WHILE{$\Vert\mathcal{M'}\Vert_0>V$}
\STATE $\mathcal{M}^*,w^*,\Omega^* \gets \mathop{\arg\min} \mathcal{L}_{total}(\mathcal{M},w,\Omega;\mathcal{B},\mathcal{H},\mathcal{C},\mathcal{D})$
\COMMENT{Optimize with Eq. (\ref{total loss})}
\STATE $\mathcal{M}^{*}[i] =
\begin{cases}
    \mathcal{M}^*[i]& if \ \mathcal{M}^*[i]>th\\
    0&otherwise
\end{cases}$
\STATE $\mathcal{M},w,\Omega \gets \mathcal{M}^*,w^*,\Omega^*$
\STATE $\mathcal{M'} \gets \mathcal{M}*G$
\COMMENT{Use Gaussian kernel $G$ as low-pass filter}
\STATE $\mathcal{M'}[i]=\begin{cases}
    1& if \ \mathcal{M'}[i] >th \\ 
    0& otherwise
\end{cases}$
\ENDWHILE
\ENSURE Brain voxel mask $\mathcal{M'}$
\end{algorithmic}
% \vspace{-0.5cm}
\end{algorithm}

Our approach is similar to the sparse mask training techniques commonly used in model pruning\cite{he2017channel,wen2016learning}. However, instead of applying sparsity constraints to the weights of linear layers, we directly impose a sparse mask on the input data. This allows us to leverage the spatial locality of brain activity to further denoise the learned mask.

% By using labeled ROIs in the dataset, we can pretrain an image decoding backbone on a cross-subject dataset. This allows us to focus on optimizing the input layer for a new subject, avoiding the need to train the entire model from scratch on a small sample dataset.

\subsection{Stage2-Learning Rate Rewinding}
A recent study \cite{renda2020comparing} suggests that retraining a model with a sparse mask from scratch outperforms fine-tuning approaches. This retraining strategy, known as learning rate rewinding, leads to improved model performance. Building on this, we retrain the TROI module from scratch using the mask $\mathcal{M'}$ trained in Stage 1, which reduces the number of voxels in the fMRI input during Stage 2. 
% Additionally, we found that L1 loss negatively affects performance in retrieval tasks. Therefore, in this stage, we exclude L1 loss from model optimization to mitigate this issue.

\section{Experiments}
\label{sec:pagestyle}
 \subsection{Experiments Setting}
We conducted the training on a single machine equipped with an L40 GPU (48GB). 
Taking Subject 1 as the target subject in our experiment as an example, for the cross-subject pretraining process, we trained the model for 150 epochs using the training data from Subjects 2, 5, and 7, with annotated ROIs in the dataset serving as voxel selectors. The same procedure applies when using other subjects as the target.
% 或许介绍一下数据集中的标注ROI？
We then used Subject 1 as a new subject for small sample learning. In Stage 1, with a batch size of 24, we reduced the number of voxels in the mask to approximately 3,000 in about 15 epochs. In Stage 2, we continued training for 150 epochs with a batch size of 24 before stopping.
% Due to the reduction in input layer size, the overall model parameters were decreased, resulting in minimal additional training cost for the TROI method.
\subsection{Dataset}
% 这里随便写点
We use the Natural Scenes Dataset (NSD) \cite{NSD}, which comprises 30,000 fMRI scans across 40 sessions from 8 subjects. Each image is viewed three times during data collection, with the viewings spaced apart. All images in the dataset are sourced from the Microsoft COCO dataset \cite{COCO}.
The fMRI data have a resolution of 1.8mm, and we perform voxel-wise z-score normalization and a rough segmentation of the brain (primarily the occipital lobe), resulting in data with dimensions of 60×40×40. This preprocessing step requires only a manual check of individual fMRI scans and does not demand precise expertise or significant manual labor.
The ROI collections provided in the NSD dataset were manually defined on fsaverage, with functional divisions based on anatomical structures. These include a set of ROIs corresponding to major sulci and gyri, as well as streams, a collection of ROIs reflecting large-scale divisions of the visual cortex (such as early, midventral, midlateral, etc.).
Our cross-subject backbone pretraining process and baseline MindEye2 are both based on these ROI annotations.

\subsection{Result}
\begin{table}[t]
\caption{
Evaluation result on dataset from subjects 1,2,5 and 7. 
A bolded number indicates a better result. 
($\uparrow$) indicates that a higher value is preferable, and ($\downarrow$) indicates that a lower value is better.
}
\centering
\label{result in other subject}
\resizebox{0.7\textwidth}{!}{%
\begin{tabular}{|c|cc|cccc|}
\hline
\multirow{2}{*}{metrics} & \multicolumn{2}{c|}{retrieval}    & \multicolumn{4}{c|}{reconstruction}                                 \\ \cline{2-7} 
                         & image  $\uparrow$         & brain   $\uparrow$        & PixCorr  $\uparrow$      & SSIM $\uparrow$          & Incep  $\uparrow$          & SwAV   $\downarrow$        \\ \hline
subj1                    & \textbf{94.0\%} & 77.6\%          & \textbf{0.235} & \textbf{0.428} & 83.6\%          & 0.459          \\
w TROI                   & 93.6\%          & \textbf{85.3\%} & 0.206          & 0.382          & \textbf{85.9\%}  & \textbf{0.431} \\ \hline
subj2                    & 90.5\%          & 67.2\%          & \textbf{0.200} & \textbf{0.433} & 81.9\%          & 0.467          \\
w TROI                   & \textbf{91.3\%} & \textbf{74.3\%} & 0.171          & 0.387          & \textbf{82.2\%} & \textbf{0.461} \\ \hline
subj5                    & 66.9\%          & 47.0\%          & \textbf{0.175} & \textbf{0.405} & 84.3\%          & \textbf{0.444}          \\
w TROI                   & \textbf{67.3\%} & \textbf{47.7\%} & 0.161          & 0.376          & \textbf{84.5\%}  & 0.454 \\ \hline
subj7                    & 64.4\%          & 37.8\%          & \textbf{0.170} & \textbf{0.408} & 74.9\%          & 0.504          \\
w TROI                   & \textbf{67.3\%} & \textbf{51.7\%} & 0.168          & 0.373          & \textbf{80.1\%} & \textbf{0.471} \\ \hline
\end{tabular}%
}
% \vspace{-4mm}
\end{table}

\begin{table}[t]
\centering
\caption{Quantitative comparison regarding the pretrained backbone (PT), low-pass filtering (LPF), and learning rate rewinding (LRR) techniques. 
Results are based on the first 1-hour data of Subject 1, with the ROI mask containing 3,000 voxels. 
% 可以的话写短一些
}
\label{ablation}
\resizebox{0.7\textwidth}{!}{%
\begin{tabular}{|c|cc|cccc|}
\hline
\multirow{2}{*}{metrics} & \multicolumn{2}{c|}{retrieval} & \multicolumn{4}{c|}{reconstruction} \\ \cline{2-7} 
                        & image  $\uparrow$        & brain  $\uparrow$       & PixCorr $\uparrow$ & SSIM $\uparrow$  & Incep $\uparrow$  & SwAV $\downarrow$ \\ \hline
Ours              & \textbf{93.6\%}         & \textbf{85.3\%}       & \textbf{0.206}    & 0.382  & 85.9\%  & \textbf{0.431} \\ \hline
w/o  PT          & 81.7\%         & 70.7\%        & 0.176    & \textbf{0.431} & 74.9\%  & 0.467 \\ \hline
w/o LPF      & 89.3\%         & 78.7\%        & 0.174    & 0.321  & 83.6\%  & 0.441 \\
+ w/o LRR       & 63.3\%         & 44.3\%        & 0.130    & 0.298  & 75.1\%  & 0.504 \\ \hline
\end{tabular}
}
% \vspace{-0.5cm}
\end{table}

% \begin{table}[htb]
% \centering
% \caption{Params of module}
% \label{tab:Params}
% \resizebox{0.8\columnwidth}{!}{%
% \begin{tabular}{|c|ccc|}
% \hline
% module          & \multicolumn{1}{c|}{stage1} & \multicolumn{1}{c|}{stage2} & MindEyeV2  \\ \hline
% TROI            & \multicolumn{1}{c|}{410M}   & \multicolumn{1}{c|}{25M}    & 61M  \\ \hline
% Backbone+MLP    & \multicolumn{3}{c|}{1.9B}                                        \\ \hline
% Diffusion Prior & \multicolumn{3}{c|}{260M}                                        \\ \hline
% Total           & \multicolumn{1}{c|}{2.6B}   & \multicolumn{1}{c|}{2.2B}   & 2.2B \\ \hline
% \end{tabular}%
% }
% \end{table}

\subsubsection{Evaluation Metrics}
We follow the same evaluation metrics as recent works \cite{ME1,ME2,ozcelik2022reconstruction,ozcelik2023natural,kneeland2023reconstructing}, as shown in Table \ref{ablation}. In the retrieval task, \textbf{image retrieval} refers to the percentage of correct image retrievals out of 300 candidates given the associated brain sample (chance = 0.3\%), vice-versa for \textbf{brain retrieval}. In the reconstruction task, \textbf{PixCorr} represents the pixel-wise correlation between the ground truth and reconstructions, and \textbf{SSIM} refers to the Structural Similarity Index Metric \cite{wang2004image}. \textbf{SwAV} represents the average correlation distance of features derived from SwAV-ResNet50's\cite{caron2020unsupervised} feature space, while \textbf{Incep} refers to two-way identification (chance = 50\%) using Inception V3 \cite{szegedy2016rethinking}.

\subsubsection{Comparison with baselines}
Fig. \ref{fig:recon} shows that we can accurately reconstruct visual stimuli, and we slightly outperform MindEye2 in small sample scenarios.
Table \ref{result in other subject} shows how our method performs on different subject.
Each experiment involves pretraining the backbone on datasets excluding the target subject, followed by fine-tuning the TROI using the first 1-hour data from the target subject's dataset, and finally evaluating on the full dataset of the target subject.

% Table \ref{metrics} shows the performance of different ROI mask, demonstrates that our approach outperforms the baseline model, MindEye2, particularly in the brain retrieval metrics, while utilizing one fifth voxels compared to the baseline.
\subsubsection{Visualization of trained ROI}
\begin{figure}[t]
\centering
    \includegraphics[width=1.0\linewidth]{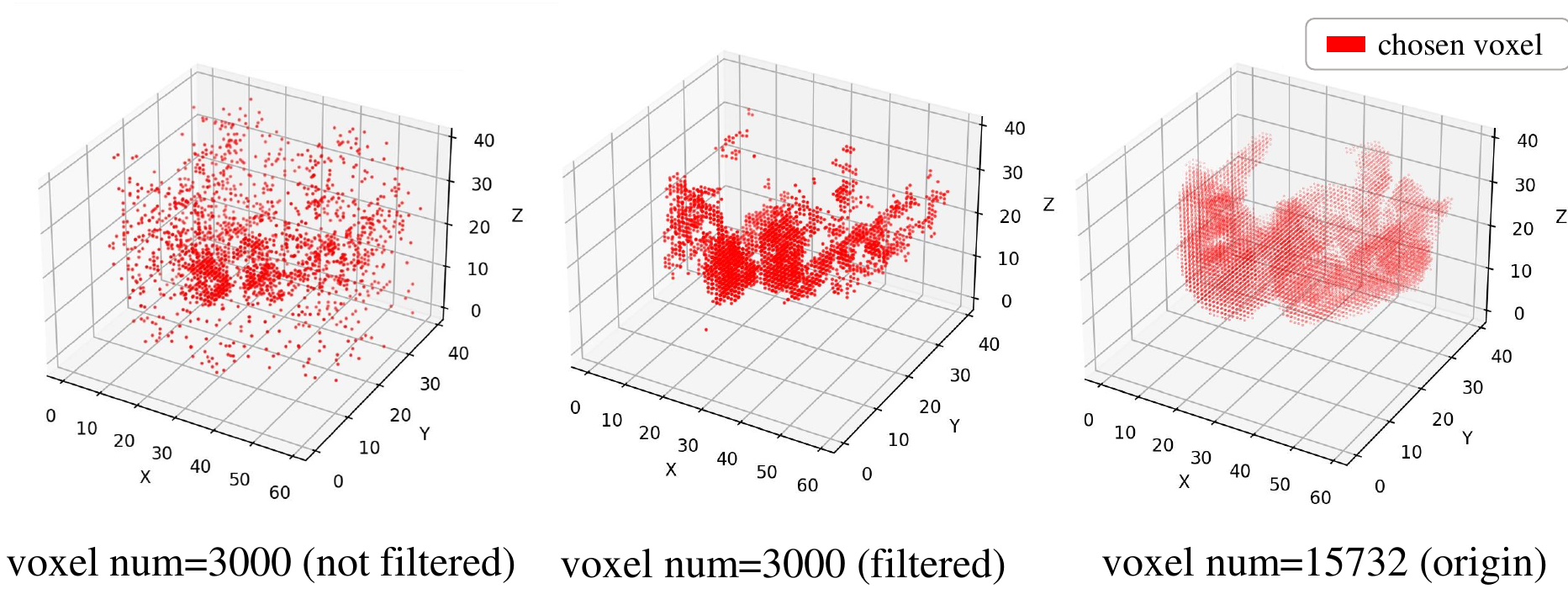}
    \caption{TROI mask trained with the first 1h data of subject 1}
    \label{fig:roi mask}
    % \vspace{-0.5cm}
\end{figure}

Fig. \ref{fig:roi mask} 
reveals that the filtered mask we trained shows a higher consistency with the annotated ROI masks in terms of spatial location compared to the unfiltered mask, which highlights the necessity of applying low-pass filtering to refine the mask, as it helps eliminate noise. The significant reduction in voxel count demonstrates our method's ability to effectively select vision-related voxels from coarsely segmented brain region aligning with findings from prior neuroscience studies\cite{grill2004human}.
\subsection{Ablation Study}

\begin{table}[t]
\caption{Quantitative comparison $w.r.t$ different ROI mask. Results were derived from the first 1h data of Subject 1. }
\centering
\label{metrics}
\begin{tabular}{|cc|cccc|}
\hline
\multicolumn{2}{|c|}{\multirow{2}{*}{metrics}}                   & \multicolumn{4}{c|}{voxel number budget}    \\ \cline{3-6} 
\multicolumn{2}{|c|}{}                                          & 1000   & 2000   & 3000   & baseline    \\ \hline
\multicolumn{1}{|c|}{\multirow{2}{*}{retrieval}}      & image $\uparrow$    & 90.3\% & 93.7\% & 93.6\% & \textbf{94.0\%} \\
\multicolumn{1}{|c|}{}                                & brain $\uparrow$    & 76.5\% & 84.7\% & \textbf{85.3\%} & 77.6\% \\ \hline
\multicolumn{1}{|c|}{\multirow{4}{*}{recon}} & PixCorr $\uparrow$ & 0.198  & 0.213  & 0.206  & \textbf{0.235}  \\
\multicolumn{1}{|c|}{}                                & SSIM $\uparrow$   & 0.376  & 0.388  & 0.382  & \textbf{0.428}  \\
\multicolumn{1}{|c|}{}                                & Incep $\uparrow$   & 78.7\%  & 83.8\%  & \textbf{85.9\%}  & 83.6\%  \\
\multicolumn{1}{|c|}{}                                & SwAV $\downarrow$   & 0.504  & 0.477  & \textbf{0.431}  & 0.459  \\ \hline
\end{tabular}
% \vspace{-0.4cm}
\end{table}

\textbf{Impact of pretrained backbone, low-pass filtering and learning rate rewinding}
Table \ref{ablation} shows that utilizing the pretrained backbone network improves the model's performance across all metrics in the small-sample scenario, confirming the effectiveness of our pretraining approach.
Table \ref{ablation} also shows that low-pass filter and learning rate rewinding method contribute to improving the model's decoding performance from fMRI data. Although the learning rate rewinding approach incurs additional training costs, our experimental results demonstrate that it is essential for effective decoding in small-sample scenarios.

\textbf{Impact of voxel number budget}
Table \ref{metrics} shows that the decoding performance improves consistently with an increasing voxel number budget. Interestingly, the model utilizing only 3,000 voxels already outperforms the baseline model that relies on the original ROI. These results support our hypothesis that a substantial fraction of voxels within the annotated ROI contribute minimally to the decoding task.

\section{Conclusion}
\label{sec:typestyle}
This paper introduces a data-driven ROI labeling approach, termed TROI, designed to decode fMRI brain visual activity in new subjects, especially when ROI annotations are unavailable and the training data is limited. To mitigate noise, we applied the sparse mask training and learning rate rewinding method, significantly reducing the number of brain voxels in the input data. Additionally, by employing a cross-subject pretraining method, we minimized the model’s trainable parameters in small-sample fine-tuning, effectively mitigating the risk of over-fitting. Using MindEye2, the state-of-the-art model for fMRI visual decoding, as our baseline, we trained on a small sample dataset from NSD and surpass the annotated ROI’s performance, particularly in terms of brain retrieval metrics.

\section*{Acknowledgment}
This work was supported in part by National Key Research and Development Program of China under Grant No. 2020YFA0804503, National Natural Science Foundation of China under Grant No. 62272264, and Beijing Academy of Artificial Intelligence (BAAI). 

%Bibliography
\bibliographystyle{unsrt}  
\bibliography{refs}

\end{document}